# Human Emotional Facial Expression Recognition

Chendi Wang, Southeast University, China

*Abstract*—**An automatic Facial Expression Recognition (FER) model with Adaboost face detector, feature selection based on manifold learning and synergetic prototype based classifier has been proposed. Improved feature selection method and proposed classifier can achieve favorable effectiveness to performance FER in reasonable processing time.**

## I. INTRODUCTION

EMOTION intelligence has become an active topic in human–computer interaction (HCI) applications. Systems with ability to detect emotional state of human, have a wide range of applications, including medicine, security, education, psychiatry, business, etc. [1].

Psychologist indicated that facial expression (FE) is the dominant component of emotion, contributing for 55 percent of the emotional expression [2]. Also, it is widely accepted that emotions can be classified into a primary set of basic discrete emotions, which is identified with characteristic FE that had substantial cross-cultural expression and recognition [3]. As a result, tracking emotional cues through FER is important for improving the performance of HCI in more natural ways.

Due to its wide range of applications, automatic FER has attracted much attention in recent years [4]. Various techniques have been developed for automatic FER, which differ in image preprocessing methods, feature extraction methods, feature selection methods and classifiers design.

*Related Work*
*Feature Extraction*

Two types of features can be extracted: geometric features and appearance features. Geometric features present the shape and locations of facial components (including mouth, eyes, brows, and nose). The appearance features present the appearance (skin texture) changes of the face, such as wrinkles, bulges and furrows.

In the geometric features framework, accurate head and face models have to be constructed manually, which is a tedious undertaking [5]. Automatic active appearance model (AAM) mapping can be employed to reduce the manual preprocessing of the geometric feature initialization [6].

On the other hand, appearance feature-based approaches allow to separate fairly well different information sources such as facial illumination and deformation changes and can be realized more automatically [6]. In the appearance features framework, methods can be categorized according to whether they act holistically or locally.

The holistic methods include PCA (principal component analysis), ICA (independent component analysis) and LDA (linear discriminant analysis). *PCA* is an unsupervised method without the requirement of class information, which can efficiently handle high dimensional data. However, the dimension of the covariance matrix is too high, which is not preferred in real-time applications. *ICA* is an unsupervised method that can extract hidden information between pixels and is suitable for non-Gaussian distribution data, while real-time implementation is still a problem. FastICA has been proposed to achieve a fast feature extraction. *LDA* is a supervised method sufficiently retaining the class structure. And SVDA(Support Vector Discriminate Analysis) method has been proposed based on Fisher linear discriminant analysis and support vector machine(SVM) in [7].This method can generate maximum inter-class separability using small samples, resulting in higher accuracy compared to PCA and ICA.

The local methods mainly include Gabor wavelets and LBP (Local Binary Pattern) operator. Gabor wavelets are widely used to extract the facial appearance changes as a set of multiscale and multiorientation coefficients. Gabor filters remove also most of the variability in images that occur due to lighting changes. However, calculation at different scales and orientations of wavelet kernel function makes it harder for real-time applications. As a result, hybrid geometric and appearance feature-based methods has been used by generating Gabor coefficients at the fiducial points from the geometric model, which can achieve higher accuracy [8].Compared to Gabor wavelet, LBP operator can extract facial expression features more effectively. But the binary thresholding is susceptible to noise. CBP (Centralized Binary Pattern) has been proposed in [9] to reduce the dimension of the LBP histogram and enhance the robustness to noise with higher accuracy by modifying the sign function.

*Classification*

The final step of FER systems is to recognize facial expression based on the extracted features. Many classifiers have been used such as neural network (NN), support vector

Chendi Wang is with Biomedical Engineering Department, Southeast University, China (e-mail: chendi.wang.judy@ gmail.com).

machines (SVM), linear discriminant analysis (LDA), K-nearest neighbor(KNN), Gaussian mixture models (GMMs), logistic regression (LR), hidden Markov models (HMM), naive Bayes(NB), tree augmented naive Bayes(TAN), and others.

NN model is hard to build and easy to fail when it applied to testing samples. HMM is usually designed for sequence-based FER incorporating motion information from successive frames. Classifiers based on Bayes have strict constrains of distribution of features, which makes the assumption that all features are conditionally independent. SVM, which is developed based on the structural risk minimization principle from statistical learning theory, has some limitations when it applied to multi-class problems generating ambiguous area in classification. KNN is distance-based method, which is simple and robust, but slow when there are many attributes and it needs to keep all the instances in the memory. Among these classifiers, the best classification results are obtained with the KNN classifier [1]. It seems that FER is not a simple classification problem and all the models tried (e.g., NB, TAN, SVM) were not able to entirely capture the complex decision boundary that separates the different facial expressions.

*Short Summary of submitted work*

Although much progress has been made, FER with a high accuracy is still difficult due to the subtlety, complexity and variability of facial expressions [4]. In this paper, FER from static images will be particularly studied with emphasis on feature extraction (Gabor scales selection & improved LBP method CBP), feature selection (proposed Manifold learning based method) and classifier design (proposed prototype based nearest neighbor classifier). The experimental results indicate that this FER model can improve the accuracy within a reasonable processing time, which will be quite useful for emotion studies and HCI applications.

## II. MATERIAL

The model will automatically detect frontal faces and perform FER on 7 emotional states: neutral state, happiness, sadness, fear, surprise, anger and disgust. Pictures of Facial Affect (POFA) FE database [10], developed by Dr. Paul Ekman and Wallace Friesen, Human Interaction Laboratory, University of California Medical Center, San Francisco has been used. POFA dataset collection consists of 110 photographs of 14 subjects that have been widely used in cross-cultural studies and neuropsychological research.

## III. METHODS

Generic facial expression recognition framework can be distinguished by certain sections: image preprocessing, face detection, face normalization, facial feature extraction, feature selection and classification.

### A. Image Preprocessing

Denoising-Enhancement: There will be noise and limited contrast in the images, which requires denoising and enhancement before further processing. In this paper, mean filter has been used to noise removal and Histogram Equalization has been used to performance Greyscale contrast enhancement in Fig. 1.

### B. Face Detection

Accurate and fast detection of region of interest, i.e. face in the background images is the vital step in FER. There are various methods for human face detection (template based, NN, SVM, etc.). Among them, Adaboost is a method that is capable of processing images extremely rapidly while achieving high detection accuracy. Adaboost, short for Adaptive Boosting, is a machine learning algorithm to build a strong classifier from certain week classifiers. Face detection based on Adaboost combines the Haar 2D features, "integral images" and "cascade" to obtain high face detection rates [11]. The detected face is indicated in the rectangular in Fig. 2. The accuracy of face detection in the work for POFA dataset is 100%.

### C. Face Normalization

Since face images are shot in different environments and times, they are not identical in intensity or facial configuration, normalization will be necessary for further feature extraxtion.

*Intensity*: Homomorphic filtering can be used to remove multiplicative noise induced by non-uniform illumination on the different areas on face. The normalized face with uniform intensity is shown in Fig. 3.

*Geometry*: To extract Gabor or LBP features, the same dimension and corresponding points of the feature are required. However, rotation of face exists sometimes and the scales of face size are always various. The solution is to identify a reference line, whose length and the angle between the horizontal axis can be used to generate images with uniform size (120 by 120 pixel) and direction. Two matching points at the end of reference line can be identified by integral projection curve and Harris corner points, which is shown in Fig. 4. The details can be found in Appendix.

### D. Facial Feature Extraction

#### 1) Gabor wavelet filter banks

Gabor filter banks are reasonable models of visual processing in primary visual cortex and are one of the most successful approaches for processing images of the human face [8]. Except for robustness to lighting changes and trait of multiresolution analysis, Gabor method is also a good way to extract the subtle changes in facial expressions, since mammalian visual cortex simple cells can be simulated using Gabor function, which means that Gabor transform is similar to perception of human visual system. To extract information about facial expression, each 120 by 120 pixel image, $I$, was convolved with a multiple spatial resolution, multiple orientation set of 2D Gabor kernels defined in (1).

$$\psi_{u,v}(z) = \frac{\left\|k_{u,v}\right\|^2}{\sigma^2} \exp\left(-\frac{\left\|k_{u,v}\right\|^2 \left\|z\right\|^2}{\sigma^2}\right)[\exp(ik_{u,v}z) - \exp(-\frac{\sigma^2}{2})] \quad (1)$$

Amongst, $u$ and $v$ denote the Gabor kernels in different orientations and frequencies. Generally, 40 kernels will be generated by $u = 0, 1, 2, 3, 4, 5, 6, 7$ ; $v = 0, 1, 2, 3, 4$ . $z = (x, y)$ means the coordinate position of a pixel in the original image $I$, and $\|\bullet\|$ denotes a modulo operation. $k_{u,v} = k_v e^{i\phi^u}$ is the center frequency of the Gabor kernel , where $k_v = k_{max} / \lambda^v$ , $\phi_u = \pi u / 8$ , $\lambda$ is the scaling factor, generally $k_{max} = \pi / 2$ , $\lambda = \sqrt{2}$ , $\sigma = \pi$ . Meanwhile, $\frac{\|k_{u,v}\|^2}{\sigma^2}$ represent the compensation of energy spectrum attenuation in different frequency bands, $\exp(-\frac{\|k_{u,v}\|^2 \|z\|^2}{\sigma^2})$ is the Gaussian convolution function .

The Gabor wavelet representation is essentially the concatenated pixels of the 40 modulus-of-convolution images obtained by convolving the input image $I = I(x, y)$ with these 40 Gabor kernels in (1): $G_{u,v} = I(x, y) * \psi_{u,v}(x, y)$ , which are shown in Fig. 5. Seen from the results, the local energy spectrum energy distributes mainly around areas of mouth, eyes, eyebrows, which differ greatly to represent various emotional states.

*Size of Gaussian Convolution Template(GCT)*

Size of GCT also affects results of Gabor Transformation. Essentially, small size template indicate global feature while large size template results in local details. The optimal size of 21 by 21 GCT has been chosen with the highest accuracy when it applied to image size of 120 by 120.

*Customized Scale Choice for FE*

Recent studies indicated that information of FE focus on small scales of Gabor filters rather that large ones and there is more redundant information between adjacent scales. By selecting interval scales in small scales (choose $v = 0, 2$ instead of $v = 0, 1, 2, 3, 4$ ), better recognition results can be achieved in this paper. And the dimensionality reduction has been achieved to increase the system speed at the same time.

*Downsampling*

The feature dimension after Gabor transformation is 57600 ($120 \times 120 \times 40$), which will affect classification and slow down detection speed greatly with much redundant information. In this paper, before concatenation, each output image is downsampled according to the spatial frequency of its Gabor kernel and normalized to zero mean and unit variance. The optimal downsampling size of 30 by 30 has been found with the highest accuracy.

### 2) Local Binary Pattern (LBP)

On the other hand, LBP features, as another descriptor that captures small texture details, have properties of tolerance against illumination changes and their computational simplicity[4]. Ojala et al. [12] introduced LBP operator in 1996, which takes a local neighbourhood around each pixel, thresholds the pixels of the neighbourhood at the value of the central pixel and uses the resulting binary-valued image patch as a local image descriptor.

It was defined for 3×3 neighbourhoods, giving 8 bit codes based on the 8 pixels around the central one. The LBP operator can be defined as:

$$LBP(x_c, y_c) = \sum_{n=0}^{7} 2^n s(i_n - i_c)$$
$$s(x) = \begin{cases} 1, if \ x > 0 \\ 0, if \ x \le 0 \end{cases} \tag{2}$$

where in this case $n$ runs over the 8 neighbours of the central pixel $c$, $i_c$ and $i_n$ are the gray-level values at $c$ and $n$. Since the binary data produced by LBP are sensitive to noise, CBP (Centralized Binary Pattern)[9] with modified sign function is used in this paper. In order to decrease the white noise's effect on images, sign function s(x) has been modified the to be the following form:

$$s(x) = \begin{cases} 1, if \ |x| > C \\ 0, if \ |x| \le C \end{cases} \tag{3}$$

The results are shown in Fig. 6. These two feature spaces has been applied and compared in this paper.

### E. Feature Selection

The dimension of both Gabor wavelet and LBP features is 14400, which is far more than the intrinsic dimensionality of the FE space. Classical approaches for dimensionality reduction include PCA, LDA and multidimensional scaling (MDS). In the linear setting, each variable is assumed to be linearly independent, making it problematic to nonlinear distributions. Various methods that generate nonlinear maps have also been considered. Most of them, such as self-organizing maps and other neural network–based approaches, set up a nonlinear optimization problem whose solution is typically obtained by gradient descent that is guaranteed only to produce a local optimum; global optima are difficult to attain by efficient means.

#### 1) Laplacian Eigenmaps

In the FER, faces formed by the different light conditions, posture effected by non-linear change can be defined as a low-dimensional manifold embedded in high-dimensional space. Manifold learning method is a non-parametric method of high efficiency without iterations. Laplacian Eigenmaps (LE) is one of the Manifold learning method. Based on the correspondence between the graph Laplacian, the Laplace Beltrami operator on the manifold and the connections to the heat equation, LE is a geometrically motivated algorithm for representing the highdimensional data. LE provides a computationally efficient approach(only local computations plus one sparse eigenvalue problem) to nonlinear dimensionality reduction that has locality-preserving properties(reletively insensitive to noise) and a natural connection to clustering (suitable for expression classification).

The justification for LE [13] comes from the role of the Laplace Beltrami operator in providing an optimal embedding for the manifold. The manifold is approximated by the adjacency graph computed from the data points. The Laplace Beltrami operator is approximated by the weighted Laplacian of the adjacency graph with weights chosen appropriately. The key role of the Laplace Beltrami operator in the heat equation enables us to use the heat kernel to choose the weight decay function in a principled manner. Thus, the embedding maps for the data approximate the eigenmaps of the Laplace Beltrami operator, which are maps intrinsically defined on the entire manifold.

*LE algorithm*

Given $k$ points $x_0,...,x_{k-1}$, expressed as set $X$ in $R^l$, we construct a weighted graph with $k$ nodes, one for each point, and a set of edges connecting neighboring points. The embedding map $y_0,...,y_{k-1}$ in $R^m$, $m << l$, expressed as set $Y$ is now provided by computing the eigenvectors of the graph Laplacian. The algorithmic procedure is formally stated below.

Step 1: Constructing the adjacency graph. We can put an edge between nodes $i$ and $j$ if $x_i$ and $x_j$ are "close." There are two variations:

(a) ε-neighborhoods (parameter $\varepsilon \in R$). Nodes $i$ and $j$ are connected by an edge if $\left\| x_i - x_j \right\|^2 < \varepsilon$ where the norm is the usual Euclidean norm in $R^l$. *Advantages*: Geometrically motivated, the relationship is naturally symmetric. *Disadvantages*: Often leads to graphs with several connected components, difficult to choose ε.

(b) $n$ nearest neighbors (parameter $n \in N$). Nodes $i$ and $j$ are connected by an edge if $i$ is among $n$ nearest neighbors of $j$ or $j$ is among $n$ nearest neighbors of $i$. Note that this relation is symmetric. *Advantages*: Easier to choose; does not tend to lead to disconnected graphs. *Disadvantages*: Less geometrically intuitive. This criteria has been used in this paper.

Step 2: Choosing the weights. There are two variations for weighting the edges:

(a) Heat kernel (parameter $t \in R$). If nodes $i$ and $j$ are connected: $W_{ij} = e^{\frac{\left\| x_i - x_j \right\|^2}{t}}$, otherwise, put $W_{ij} = 0$.

(b) Simple-minded (no parameters ($t = \infty$)). $W_{ij} = 1$ if vertices $i$ and $j$ are connected by an edge and $W_{ij} = 0$ if vertices $i$ and $j$ are not connected by an edge. This simplification avoids the need to choose $t$.

Step 3: Eigenmaps. Assume the graph $G$, constructed above, is connected. Otherwise, proceed with step 3 for each connected component. Compute eigenvalues and eigenvectors for the generalized eigenvector problem,

$$Ly = \lambda Dy \qquad (4)$$

where $D$ is diagonal weight matrix, and its entries are column (or row, since $W$ is symmetric) sums of $W$, $D_{ii} = \sum_j W_{ji}$.

$L = D - W$ is the Laplacian matrix. Laplacian is a symmetric, positive semidefinite matrix that can be thought of as an operator on functions defined on vertices of $G$.

Let $y_0,...y_{k-1}$ be the solutions of (4), ordered according to their eigenvalues:

$$Ly_0 = \lambda_0 Dy_0$$
$$Ly_1 = \lambda_1 Dy_1$$
$$\cdots$$
$$Ly_{k-1} = \lambda_{k-1} Dy_{k-1}$$
$$\lambda_0 \leq \lambda_1 \leq \cdots \leq \lambda_{k-1}$$

The performance of LE has been compared with FER accuracy before Dimensionality Reduction(DR). Seen from the results in table I, the heat kernel weighting calculation is better than simple minded since it can reflect the distance information between two neighboring points. At the same time, LE has a slight increase of accuracy compared to results before DR and it still needs to be improved.

*2) Supervised Divergence Laplacian Eigenmap(SDLE)*

There are limitations in LE algorithm, which prevent the favorable improvement of FER. (a) LE only guarantees the Neighborhood trait of the samples in the embedded space, however, it does not consider the relationship of non-neighbors. In this case, samples far away from each other in manifold may be mapped to neighborhood in embedded space, which will affect the result of FER. (b) There are multiple categories (sub-manifolds) in the context of FER. Then, those close samples belonging to different categories will impact the performance of mapping matrix. SDLE has been proposed to solve these two problems in this paper.

On one hand, SDLE tries to ensure the submanifold structure to be local invariant by representing both neighbor and non-neighbor traits; on the other hand, it makes dispersed projection of samples in different categories based on differences in submanifolds. SDLE is realized based on intra and inter class Divergence.

*Intra Divergence $J_L$*

One goal of SDLE is to scatter the points in the embedded space which are far from each other in the original sample space, while ensure that the points close to each other in original sample space are still neighbors in the embedded space. Intra-class Divergence function $S_D$ in (5) is proposed, which defines intra-class Divergence $J_L$ by Euclidean distance measurements with supervised information. Euclidean distance can be replaced by similarity here, which can achieve better performance of FER.(the closer two points are, the smaller Euclidean distance is and the higher the similarity is).

$$S_D = \begin{cases} simi(x, y) = sigmoid(simi(x, y), p, a), & if \ lable(x) = lable(y) \\ simi(x, y) = simi(x, y) - \min \ simi(x, y), & if \ lable(x) \neq lable(y) \end{cases} \quad (5)$$

Where $sigmoid(z, p, a) = \dfrac{1}{1 + e^{-p(z-a)}}$ and $z = simi(x, y)$.

The regular similarity of two close points in one category is larger than 0.5, then the first equation of $S_D$ will be applied to prize the similarity by the right half of sigmoid in Fig. 7. If the similarity in this case is less than 0.5, indicating that noise or outliers happen, then the points certainly can not be put into neighbor area. As a result, the left half of sigmoid is use to punish the corresponding similarity. Meanwhile, there will be points from different categories but with high similarity(close distance), which should be punished and suppressed by subtracting a fixed value.The new weighting matrix can be defined as: $W_{ij}^{s_D} = \exp(-\frac{S_D(simi(i,j))}{t})$ , then the intra-class Divergence $J_L$ is as follows:

$$
\begin{aligned}
J_L &= \sum_{ij} \left\| Simi(Y_i, Y_j) \right\| \cdot W_{ij}^{s_D} \\
&= 2\sum_i Y_i D_{ii} Y_i^T - 2\sum_{ij} Y_i \cdot W_{ij}^{s_D} \cdot Y_j^T \\
&= 2tr\left\{ Y(D - W^{S_D})Y^T \right\}
\end{aligned}
\tag{6}
$$

$where\ D_{ii} = \sum_j W_{ij}^{S_D}.$

*Inter Divergence $J_M$*

Distance between different sub-manifolds must be expanded. Matrix $M$ can be used to represent the class information of any two points. As a result, the calculation of Inter Class Divergence $J_M$ is based on $M$.

$$
M_{ij} = \begin{cases} 0 & if\ lable(i) = lable(j) \\ 1 & if\ lable(i) \neq lable(j) \end{cases}
$$

$$
\begin{aligned}
J_M &= \sum_{ij} M_{ij} \left( \left\| Simi(Y_i, Y_j) \right\| \right) \\
&= 2\sum_i Y_i P_i Y_i^T - 2\sum_{ij} Y_i \cdot M_{ij} \cdot Y_j^T \\
&= 2tr\left\{ Y(P - M)Y^T \right\}
\end{aligned}
\tag{7}
$$

$where\ P_{ii} = \sum_j M_{ij}.$

To obtain the mapping transformation matrix, which is needed when applied to new testing points, (6) and (7) can be rewritten as:

$$
J_L = 2tr\left\{ Y(D - W^{S_D})Y^T \right\} = 2tr\left\{ A^T X(D - W^{S_D})X^T A \right\}
\tag{8}
$$

$$
J_M = 2tr\left\{ Y(P - M)Y^T \right\} = 2tr\left\{ A^T X(P - M)X^T A \right\}
\tag{9}
$$

*Objective Function*

The goal of SDLE is to extract the most appropriate classification features without destroying the topological structure of sub-manifolds. In the mathematical expression, SDLE needs to achieve the objective function:

$$
\begin{cases} J(A) = \max(J_M) = tr\left\{ A^T X(P - M)X^T A \right\} \\ under condition: tr\left\{ A^T X(D - W^{S_D})X^T A \right\} \end{cases}
\tag{10}
$$

To solve this, the constraint and objective function can be combined into a linear functional:

$$
\begin{aligned}
\tilde{J}(A) &= \max\left\{ J_M - \lambda tr\left( A^T X(D - W^{S_D})X^T A - X(D - W^{S_D})X^T \right) \right\} \\
&= \max\left\{ A^T X(P - M)X^T A - \lambda\left( A^T X(D - W^{S_D})X^T A - X(D - W^{S_D})X^T \right) \right\} \\
&= \max\left\{ A^T X(P - M)X^T A - \lambda A^T X(D - W^{S_D})X^T A + \lambda X(D - W^{S_D})X^T \right\}
\end{aligned}
\tag{11}
$$

By finding the extrema A of this functional:

$$
\frac{\partial \tilde{J}(A)}{\partial A} = 2X(P - M)X^T A - 2\lambda X(D - W^{S_D})X^T A = 0
\tag{12}
$$

the mapping feature sapce $Y$ can be found by: $Y = A^T X$ .

The final feature dimension after SDLE dropped from 14400 to 83, with the FER accuracy in table II. Seen from the results, SDLE has a great increase in accuracy compared to pre-DR and DR using LE.

*F. Classification*

The distance-based classifier KNN has been proven to be better at FER than bayesian networks, decision trees, SVM, etc. [1]. Although KNN may give better classification results but it has its own disadvantages: it is computationally slow and needs to keep all the instances in the memory. To solve this problem, it has been considered to build a classifier that combines the synergetic prototype [14] with KNN to better represent the subclass without storing all the training samples. There are two steps in this classifier:

(a) Prototypes Generation: information of training samples can be integrated to get representative prototypes of different classes.

(b) Nearest Neighborhood Classifier: Compute the similarity of test sample and the prototypes of different classes, then output is chosen by the prototype with maximal similarity.

There are three approaches to generate the prototypes, which are MEAN, CLUSTER and FEEDBACK respectively. MEAN is to compute the arithmetic means of all the training samples in one certain class. It is simple and fast, however, this method treats all the samples equally without the learning ability. CLUSTER computes the cluster centers with higher accuracy than MEAN in a much slower way. FEEDBACK is improved based on MEAN incorporating error information. It can achieve high accuracy and is much faster than CLUSTER. As a result, FEEDBACK combined with SDLE and Similarity measurement with highest accuracy of 93.29% has been used in this paper, as shown in table III.

## IV. RESULTS

The result of image preprocessing is shown in Fig. 1. Images with less noise higher contrast can be obtained. Fig. 2 shows the detected faces using Adaboost with accuracy of 100% in this POFA dataset. Fig.3 and 4 shows the results of Intensity and Geometric normalization respectively. We can get images with size of 120 by 120 pixel and uniform horizontal angle. The features extracted based on Gabor wavelets and LBP operator are shown in Fig. 5 and 6 respectively. And Fig. 7 is the illustration of the Intra-class Divergence function $S_D$ with supervised information in SDLE.

The processing time of each section of the FER system in this paper is shown in the timebar as follows(in secoend):

| Ada | Geometric | Gabor | LBP | SDLE | Classifier |
|-----|-----------|-------|-----|------|------------|
| 0.230 | 0.253 | 0.058 | 0.007 | 0.004 | 0.027 |

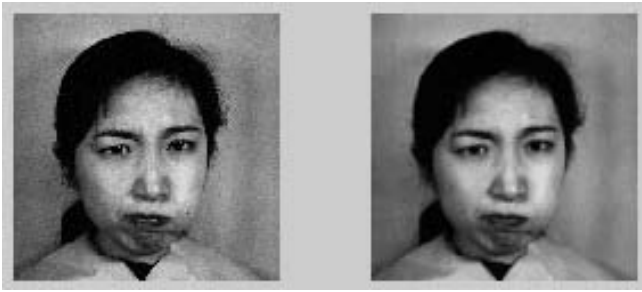

Fig. 1. Demonstration of image preprocessing. Left and right images are original image and image after preprocessing respectively.

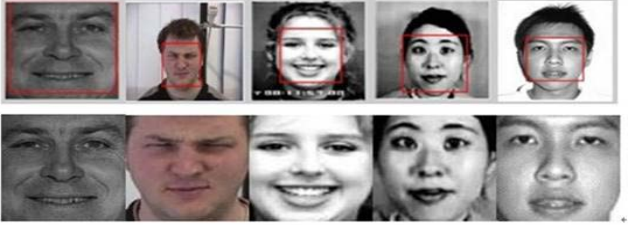

Fig. 2. Face Detection based on Adaboost method. The regions indicated by the red rectangles are the results of detected face area.

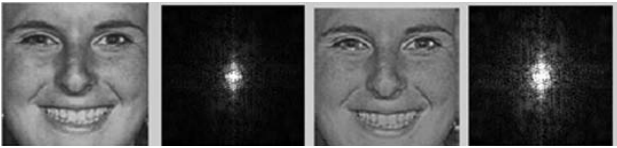

Fig. 3. Intensity normalization using Homomorphic filtering. Two images on the left is original image and its frequency domain and those on the right are result after filtering.

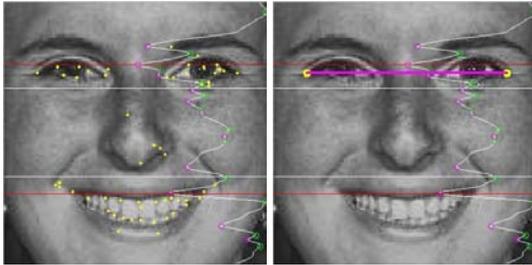

Fig. 4. Geometric normalization. Left image is the result of corner detection. The magenta line in the image on the right is the reference line used to generate uniform size and orientation.

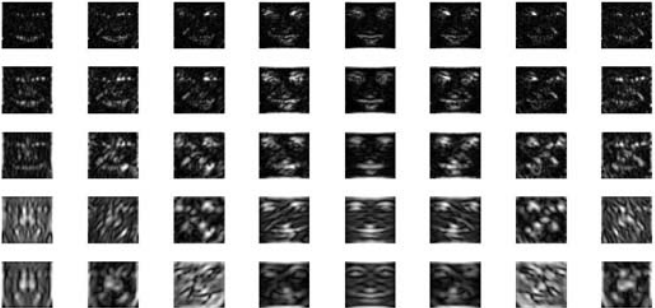

Fig. 5. Image transformed by 40 Gabor filters with 5 scales and 8 orientations.

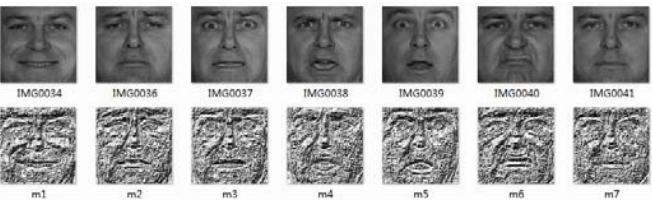

Fig. 6. Images of 7 emotion states transformed by LBP operator.

TABLE I
FER ACCURACY USING LE (%)

| Non | Simple | Heat kernel |
|---|---|---|
| 69.13 | 72.45 | 76.53 |

TABLE II
FER ACCURACY USING SDLE (%)

| Non | LE | SDLE |
|---|---|---|
| 69.13 | 76.53 | 88.63 |

TABLE III
FER ACCURACY USING PROTOTYPE (%)

| KNN | MEAN | CLUSTER | FEEDBACK |
|---|---|---|---|
| 69.13 | 76.53 | 88.63 | 93.29 |

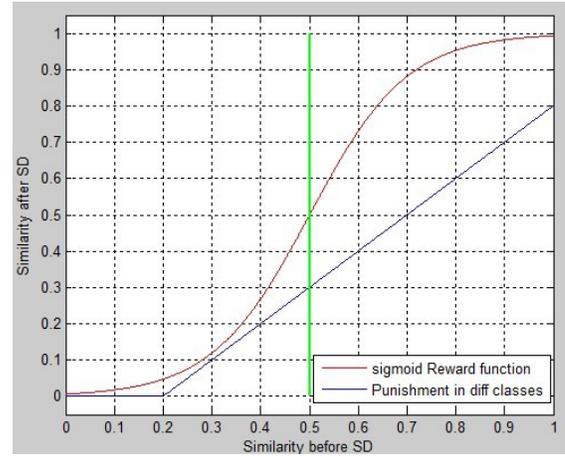

Fig. 7. Intra-class Divergence function $S_D$ in SDLE

| TAG | STAT | | | | Evaluation (%) | | |
|---|---|---|---|---|---|---|---|
| | TP | FP | FN | TN | SEN | SPE | ACE |
| 1 | 14 | 0 | 0 | 84 | 100 | 100 | 100 |
| 2 | 10 | 4 | 4 | 80 | 71.43 | 95.24 | 91.84 |
| 3 | 11 | 5 | 3 | 79 | 78.57 | 94.05 | 91.84 |
| 4 | 9 | 4 | 5 | 80 | 64.29 | 95.24 | 90.82 |
| 5 | 11 | 4 | 3 | 80 | 78.57 | 95.24 | 92.86 |
| 6 | 11 | 1 | 3 | 83 | 78.57 | 98.81 | 95.92 |
| 7 | 9 | 5 | 5 | 79 | 64.29 | 94.05 | 89.80 |
| AVR | | | | | 76.53 | 96.09 | 93.29 |

Fig. 8. Sensitivity, Specification and Accuracy of FER using SDLE and Gabor

| NUM | ① Happy | ② Sad | ③ Fear | ④ Anger | ⑤ Surp | ⑥ Disg | ⑦ Calm |
|---|---|---|---|---|---|---|---|
| ①: 14 | 14 | 0 | 0 | 0 | 0 | 0 | 0 |
| ②: 14 | 0 | 10 | 0 | 2 | 1 | 0 | 1 |
| ③: 14 | 0 | 0 | 11 | 0 | 1 | 1 | 1 |
| ④: 14 | 0 | 1 | 2 | 9 | 0 | 0 | 2 |
| ⑤: 14 | 0 | 0 | 2 | 0 | 11 | 0 | 1 |
| ⑥: 14 | 0 | 1 | 1 | 0 | 1 | 11 | 0 |
| ⑦: 14 | 0 | 2 | 0 | 2 | 1 | 0 | 9 |

Fig. 9. Confusion Matrix of FER using SDLE and Gabor

| TAG | STAT | | | | Evaluation (%) | | |
|-----|----|----|----|----|-----|-----|-----|
| | TP | FP | FN | TN | SEN | SPE | ACE |
| 1 | 12 | 3 | 2 | 81 | 85.71 | 96.43 | 94.90 |
| 2 | 7 | 3 | 7 | 81 | 50 | 96.43 | 89.80 |
| 3 | 9 | 5 | 5 | 79 | 64.29 | 94.05 | 89.80 |
| 4 | 7 | 3 | 7 | 81 | 50 | 96.43 | 89.80 |
| 5 | 8 | 4 | 6 | 80 | 57.14 | 95.24 | 89.80 |
| 6 | 9 | 9 | 5 | 75 | 64.29 | 89.29 | 85.71 |
| 7 | 7 | 11 | 7 | 73 | 50 | 86.90 | 81.63 |
| AVR | | | | | 60.20 | 93.54 | 88.78 |

Fig. 10. Sensitivity, Specification and Accuracy of FER using SDLE and LBP

| NUM | ① | ② | ③ | ④ | ⑤ | ⑥ | ⑦ |
|-----|---|---|---|---|---|---|---|
| ①: 14 | 12 | 0 | 1 | 0 | 0 | 1 | 0 |
| ②: 14 | 1 | 7 | 0 | 0 | 1 | 2 | 3 |
| ③: 14 | 2 | 0 | 9 | 0 | 0 | 1 | 2 |
| ④: 14 | 0 | 0 | 1 | 7 | 1 | 2 | 2 |
| ⑤: 14 | 0 | 1 | 1 | 1 | 8 | 1 | 2 |
| ⑥: 14 | 0 | 1 | 1 | 1 | 0 | 9 | 2 |
| ⑦: 14 | 0 | 1 | 1 | 1 | 2 | 2 | 7 |

Fig. 11. Confusion Matrix of FER using SDLE and LBP

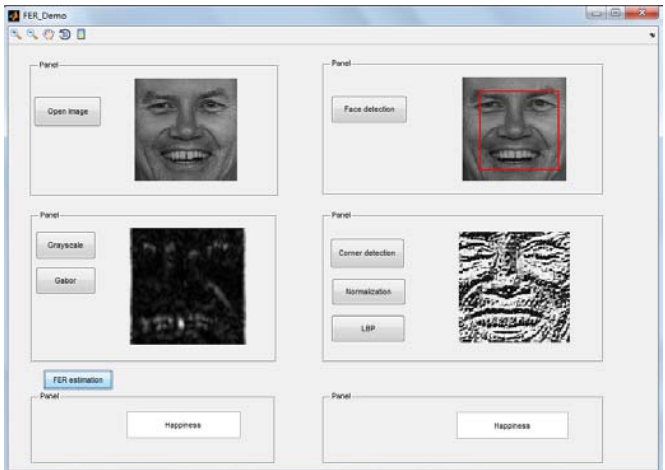

Fig. 12. FER Demo

Seen from table I, LE has a slight increase of accuracy compared to FER results before dimensionality reduction. At the same time, the heat kernel weighting calculation is better than simple minded since it can reflect the distance information between two neighboring points. Table II shows the performance of SDLE, it greatly increases the FER accuracy compared to pre-DR and DR using LE. Meanwhile, SDLE dramatically decreases the feature dimension from 14400 to 83, which is very promising for real time implementation. The comparison of different classifier has been shown in table III. The proposed classifier combined with synergetic prototype and nearest neighborhood has better FER accuracy than KNN. In the three approaches of proposed classifier, FEEDBACK prototype can achieve highest accuracy compared to MEAN, CLUSTER in rather fast speed.

Validation of this FER model with Gabor features can be reflected in Sensitivity, Specificity and Accuracy in Fig.8. Sensitivity and specificity of happiness is the highest 100%. Sensitivity of anger and calm is the lowest 64.29%, since they are difficult to be detected correctly. Specificity of both fear and surprise is the lowest. And the average accuracy of this FER model combining SDLE, synergetic prototype and similarity can get as high as 93.29%, which is very promising for accurate FER applications.

To explain the results above in a biological approach, the Confusion Matrix of FER using SDLE and Gabor is shown in Fig.9. The labels in the first row are the ground truth of emotion tags and the labels in the first column is the FER results. Seen from this confusion matrix, Happiness distinguishes well from others, since it has not been classified into other categories and there are not others being classified into happiness. Sadness is subtle and easy to be confused with clam, as a result, there are several samples which are sad being classified into calm state. Fear is confused with surprise sometimes, meaning that fear will be tagged as surprise and surprise will be tagged as fear. The reason might be that Fear and Surprise share some overlapping features such as the size of eyes and the motion of areas around mouth.

Similar results of FER model with LBP features can be found in Fig.10 and Fig. 11. FER with LBP features can achieve rather high accuracy as well, while it is lower than Gabor. As a result, Gabor wavelet has been proven to be better features for FER problem.

The final demo of the whole FER system is shown in Fig.12. This friendly GUI (Graphical user interface) can implement Human emotional facial expression recognition in reasonable processing time.

## V. CONCLUSIONS

### A. Main Contribution

FER with high accuracy and fast processing speed is challenging. There are four main contributions in this paper.

Firstly, Gabor wavelets have been chosen with the optimal combination of downsampling size, template consideration and advanced scale selection to better represent FE features.

Secondly, the proposed method SDLE proved that manifold learning can be well applied to high dimensional nonlinear feature reduction for FER. And SDLE greatly increases FER accuracy.

Thirdly, the proposed classifier combined with Nearest Neighborhood and Synergetic Prototype results in FER with higher accuracy in a faster way.

Fourthly, this FER model incorporates all the sections, including Adaboost face detection, image preprocessing, FE feature extraction, SDLE feature selection, FE classification in a complete framework, which can be implemented in reasonable processing time. This is very promising for meanful real-time FER for HCI applications.

### B. Problems Encountered

The section of geometric normalization using Harris corner is a little slow. Although this method is faster than other methods such as SIFT, there is still improvement for both the accuracy and detection speed of key points in human face. A method which can quickly detect the right key points for normalization is open to explore.

### C. Possible Future Work

First of all, except for Gabor/LBP features, other features such as geometric features can be added into feature vector. The potential combination of features of geometric based and appearance based could be a better indicator for FER.

Secondly, Advanced classifier can be designed to deal with difficult FER on tricky emotions, which share overlapping features.

Lastly, motion information extracted from video sequence should be explored to better study the underlying patterns of the transition of the change of human emotion.

## APPENDIX

*Integral Projection*

An integral projection is a one-dimensional pattern, or signal, obtained through the sum of a given set of pixels along a given direction. The horizontal and vertical integral projection are the most popular ones. Suppose $I(x, y)$ is the intensity of a pixel at location $(x, y)$, the horizontal integral projection in the $i$ th row $U(i)$ of $I(x, y)$ in intervals $[1, n]$ can be defined as:

$$U(x, y) = \sum_{y=1}^{n} I(x, y), x = 1, 2, \cdots, n$$

Since the grayscale of eyebrows, eyes , nostrils , mouth and chin is relatively lower, the corresponding horizontal integral projection point will lie in the valley of the whole curve. In order to avoid the influence of noise, the first scale approximation of wavelet transformation has been used instead of the original integral projection curve.

*Harris corner points*

The Harris corner detector[16] is a popular interest point detector due to its strong invariance to rotation, scale, illumination variation and image noise. The Harris corner detector is based on the local auto-correlation function of a signal; where the local auto-correlation function measures the local changes of the signal with patches shifted by a small amount in different directions.

Given a shift $(\Delta x, \Delta y)$ and a point $(x, y)$, the auto-correlation function is defined as:

$$c(x, y) = \sum_{W} [I(x_i, y_i) - I(x_i + \Delta x, y_i + \Delta y)]^2$$

where W is the Gaussian window centered on $(x, y)$. The shifted image is approximated by a Taylor expansion truncated to the first order terms,

$$I(x_i + \Delta x, y_i + \Delta y) \approx I(x_i, y_i) + [I_x(x_i, y_i) \; I_y(x_i, y_i)] \begin{bmatrix} \Delta x \\ \Delta y \end{bmatrix}$$

where $I_x(\cdot, \cdot)$ and $I_y(\cdot, \cdot)$ denote the partial derivatives in $x$ and $y$,

respectively. From these two equations:

$$c(x, y) = [\Delta x \; \Delta y] c(x, y) \begin{bmatrix} \Delta x \\ \Delta y \end{bmatrix}$$

where matrix $c(x,y)$ captures the intensity structure of the local neighborhood. Let $\lambda_1$, $\lambda_2$ be the eigenvalues of matrix $c(x,y)$. If both eigenvalues are high, so the local auto-correlation function is sharply peaked, then shifts in any direction will result in a significant increase; this indicates a corner.

## ACKNOWLEDGMENT

Thank you so much for Dr. Rafeef Abugharbieh, the brilliant instructor of the EECE 570 course: Fundamental of visual computing. I have acquired the knowledge of image processing in the course. Meanwhile, I have been inspired by many advanced topics in this area to better implement this challenging problem of human emotional facial expression recognition. The sections of face detection and basic KNN classifier implementation are part of my Master's work.

**Chendi Wang** received the Master degree from the Southeast University in 2013. Her research interests include Biomedical signal and image computing.